\documentclass{article}

\usepackage{PRIMEarxiv}

\errorcontextlines\maxdimen

\usepackage[utf8]{inputenc} % allow utf-8 input
\usepackage[T1]{fontenc}    % use 8-bit T1 fonts
\usepackage{url}            % simple URL typesetting
\usepackage{booktabs}       % professional-quality tables
\usepackage{nicefrac}       % compact symbols for 1/2, etc.
\usepackage{microtype}      % microtypography
\usepackage{fancyhdr}       % header
\usepackage{graphicx}
\usepackage{xcolor}
\usepackage{threeparttable}
\usepackage{booktabs}
\usepackage{multirow}
\usepackage{tabularx}
\usepackage{arydshln}

\usepackage{lineno,hyperref}
\hypersetup{colorlinks, citecolor=blue, linkcolor=red, urlcolor=green}
\usepackage{amsmath,amsfonts,amssymb}
\usepackage{graphics}
\usepackage{bm}
\usepackage{multirow}
\usepackage{algorithm}
\usepackage{algpseudocode}

\usepackage{mathtools}
\usepackage{siunitx}
\DeclarePairedDelimiterXPP\BigOSI[2]%
  {\mathcal{O}}{(}{)}{}%
  {\SI{#1}{#2}}

\title{Neural Operator induced Gaussian Process framework for probabilistic solution of parametric partial differential equations}

\author{Sawan Kumar \\
  Department of Applied Mechanics\\
  Indian Institute of Technology Delhi\\
  \texttt{sawan.kumar@am.iitd.ac.in} \\
  \And
  Rajdip Nayek\\
  Department of Applied Mechanics\\
  Indian Institute of Technology Delhi\\
  \texttt{rajdipn@am.iitd.ac.in} \\
  \And
  Souvik Chakraborty \\
  Department of Applied Mechanics\\
  Yardi School of Artificial Intelligence (ScAI)\\
  Indian Institute of Technology Delhi\\
  \texttt{souvik@am.iitd.ac.in} \\
}

\begin{document}
\maketitle

\begin{abstract}
The study of neural operators has paved the way for the development of efficient approaches for solving partial differential equations (PDEs) compared with traditional methods. However, most of the existing neural operators lack the capability to provide uncertainty measures for their predictions, a crucial aspect, especially in data-driven scenarios with limited available data. In this work, we propose a novel Neural Operator-induced Gaussian Process (NOGaP), which exploits the probabilistic characteristics of Gaussian Processes (GPs) while leveraging the learning prowess of operator learning. The proposed framework leads to improved prediction accuracy and offers a quantifiable measure of uncertainty. The proposed framework is extensively evaluated through experiments on various PDE examples, including Burger's equation, Darcy flow, non-homogeneous Poisson, and wave-advection equations. Furthermore, a comparative study with state-of-the-art operator learning algorithms is presented to highlight the advantages of NOGaP. The results demonstrate superior accuracy and expected uncertainty characteristics, suggesting the promising potential of the proposed framework. 
\end{abstract}

\keywords{Gaussian Process, neural operators, uncertainty quantification, probabilistic machine learning.}

\section{Introduction}
Neural Operators (NO)\cite{li2020fourier,lu2019deeponet,tripura2023wavelet} have recently emerged as a viable alternative to traditional numerical methods for solving partial differential equations. Some recent applications of NO in computational mechanics include weather modeling \cite{pathak2022fourcastnet,lin2023spherical}, medical diagnosis using elastography \cite{tripura2023wavelet}, fracture propagation \cite{Goswami_2022}, and fault diagnosis \cite{rani2023fault}, among others. However, a major bottleneck in the widespread adoption of NO in critical engineering systems (e.g., nuclear reactors) resides in the fact that they fail to quantify the uncertainty in the trained model. This has been recognized by the research community, and some efforts in this direction can be found in \cite{garg2023vb, Magnani2022ApproximateBN}.
An alternative to the neural operators is to employ probabilistic machine learning algorithms such as the Gaussian Process (GP) for solving partial differential equations. Unlike NO, GP is inherently probabilistic and hence, can seamlessly quantify the uncertainty in the model. Recent developments on the use of GP for solving PDEs include \cite{chen2021solving, batlle2023kernel, Chen2023SparseCF}. Unfortunately, the developments on GP and NO are carried out somewhat in parallel. To that end, the objective of this paper is to investigate how NO and GP can benefit from each other, and if a fusion of the two can result in a superior model. 

% Neural Operators and their role in PDE solving 
Neural operators learn and map two infinite-dimensional function spaces. Some neural operators that have been studied in the past and gained popularity are Wavelet Neural Operator (WNO) \cite{tripura2023wavelet}, Physics Informed WNO \cite{n2023physics}, DeepONet \cite{lu2019deeponet,lu2021learning, wang2021learning}, Fourier Neural Operator (FNO) \cite{li2020fourier}, PCA-Net \cite{hesthaven2018non, bhattacharya2021model}, Graph Neural Operators (GNO) \cite{li2020neural} and many more. Compared to classical PDE solvers, neural operators have significant computational advantages; once trained, they can make predictions for different initial and boundary conditions efficiently. The performance of both WNO and FNO is promising when compared with the state-of-the-art operator learning methods in terms of prediction accuracy. However, FNO utilizes the Fast Fourier Transform (FFT), resulting in a loss of spatial information. Consequently, frequency-localized basis functions in FFT pose challenges in learning complex domain problems. To address this variants such as geometric-FNO \cite{li2022fourier} and geometry-informed neural operators \cite{li2024geometry} have been proposed.
Alternatively, WNO utilizes wavelet transform which retains space-frequency information and can handle signals with spikes and discontinuity, thus enabling superior learning of the patterns in images. Some recent improvements of the classical WNO include \cite{tripura2023foundational,thakur2023multifidelity,rani2024generative,navaneeth2023waveformer,n2023physics}. However, as previously stated, both WNO and FNO are deterministic and cannot yield the predictive uncertainty that is essential for decision-making in safety-critical systems. 

% Application of Gaussian Processes (GPs)  
Data in real-world scenarios are accompanied by inherent noise and sources of uncertainty. These sources include errors stemming from the limitation of precision of the measuring instrument, incorrect measurement of observations, missing data, etc. Additionally, models may be built upon simplifying assumptions that may not fully capture the complexity of the underlying phenomena, which also contributes to the uncertainty. This is often addressed using probabilistic frameworks like Gaussian Process Regression (GPR) \cite{langehegermann2019algorithmic, swiler2020survey,gulian2022gaussian} and Bayesian neural operator frameworks \cite{Yang_2021, garg2022variational, zou2022neuraluq}. GPR is a non-parametric machine learning model, and by design, learns distribution over the functions. Similar to neural networks and neural operators, GPR is also a popular choice among researchers for solving partial differential equations. This includes the development of physics-informed GPR for linear and nonlinear systems \cite{pförtner2023physicsinformed,chen2023sparse,besginow2022constraining}. However, GPR suffers from the curse of dimensionality and does not scale with an increase in training samples; hence, GPR is less popular as compared to neural network-based approaches.

% Objective of the Study: Introducing NOGaP Framework 
The developments on NO and GP are mostly carried out in parallel. To that end,
the objective of this paper is to propose a novel architecture referred to as the Neural Operator induced Gaussian Process (NOGaP) that attempts to bridge the gap between the NO and GP-based approaches. NOGaP ingeniously combines the strengths of Gaussian process regression with NO. Specifically, the high-level idea is to use NO as a mean function within the GP framework. We hypothesize that with NO as the mean function, NOGaP will be able to exploit the accuracy of NO and the probabilistic nature of the GP. Without loss of generality, we use the recently proposed WNO as the mean function with the proposed approach. Additionally, to address the challenge associated with the scalability of GP, we propose to use the Kronecker -product-based approach proposed in \cite{bilionis2013multi,bonilla2007multi}.

The proposed framework has the following key features: 
\begin{itemize} 
     \item \textbf{Accuracy}:
     NOGaP, a fusion of NO and GP, is anticipated to offer superior accuracy compared to both NO and GP individually. This has been empirically illustrated in the numerical section.  
     \item \textbf{In-built uncertainty quantification capability}: NOGaP possesses built-in probabilistic capabilities, allowing it to assess the epistemic uncertainty due to limited and noisy datasets. This feature is crucial, especially for ensuring the reliability of the model in critical engineering systems. 
     \item \textbf{Scalability}: Leveraging the Kronecker product property, NOGaP exhibits scalability, enabling its efficient handling of large datasets. This marks a departure from conventional GPs, which often encounter scalability issues due to the curse of dimensionality.
%
    % \item The proposed framework exploits WNO as the mean function of the GP.
    % \item NOGaP can handle a family of highly nonlinear PDEs.
    % \item It utilizes the probabilistic characteristic of GP to give a measure of uncertainty for its prediction. 
    % \item A noteworthy feature of the framework NOGaP is enhanced prediction capabilities compared to standalone WNO. This makes the proposed framework a suitable choice for a wide range of applications. 
\end{itemize} 

 % Write the outlines of the rest of the paper 
The remainder of the paper is arranged as follows: Section \ref{sec:NOGaP} elaborates on the mathematical formulation of the proposed framework. Section \ref{sec:numerical} explores the different numerical examples on which the NOGaP is tested along with the results. Finally, the paper is concluded in the last section \ref{sec: conclusion} with results, observations and future work. 

\section{Neural Operator induced Gaussian Process (NOGaP)}\label{sec:NOGaP}
Let us define the hypothesis space, where $\mathcal{Z} \subset \mathbb{R}^d$ represents the input space and $\mathcal{Y} \subset \mathbb{R}^o$ denotes the output space. Suppose we have $N$ distinct pairs of observations with inputs, $\mathbf{Z} = (\bm{x}_1, \ldots, \bm{x}_N,\bm{a}_1, \ldots, \bm{a}_N)$, and a corresponding set of outputs, $\mathbf{Y} = (\bm{y}_1, \ldots, \bm{y}_N)$. The model takes the form: 
\begin{equation}\label{eq:mapping}  
\begin{aligned}
\bm{y}(\bm{z}) & = \bm{f}_o(\bm{z}) + \bm{\epsilon} \\
\bm{\epsilon} & \sim \mathcal{N}(\mathbf{0}, \sigma_o^2 \mathbf{I})
\end{aligned}
\end{equation}
where $\mathbf{y}$ are the noisy observations and $\bm{\epsilon} \in \mathbb{R}^o$ is modeled as independent zero-mean Gaussian white noise with variance $\sigma_o^2$. The vector-valued function $\bm{f}_o$ is assumed to follow a GP prior:
\begin{equation}
\begin{aligned}
    \bm{f}_o(\bm{z}) \sim \mathcal{GP}(\bm{m}_o(\bm{z}; \bm \theta_m), \textbf{K}_o(\bm{z}, \bm{z}^{\prime}; \bm \theta_k))
\end{aligned}
\end{equation}
Here we have \emph{vector-valued} mean function $\bm{m}_o: \mathbb{R}^d \mapsto \mathbb{R}^o$ and a \emph{matrix-valued} covariance function $\mathbf{K}_o: \mathcal{Z} \times \mathcal{Z} \mapsto \mathbb{R}^{o \times o}$, with vector-valued parameters $\bm \theta_m$ and $\bm \theta_k$ where $\bm \theta_m$ represents parameters of the mean function and $\bm \theta_k$ denotes kernel parameters.
Here in our formulation, the multioutput kernel $\textbf{K}_o$ is assumed to be separable \cite{alvarez2012kernels}. It can be expressed as:
\begin{equation}
\begin{aligned}
     (\mathbf{K}_o)_{dd'} &= k_x(\bm{z}, \bm{z'}) \cdot k_f(d,d') \\
     \mathbf{K}_o &= \mathbf{K}_x \otimes \mathbf{K}_f
\end{aligned}
\end{equation}
where \(\otimes\) denotes the Kronecker product, $\mathbf{K}_o$ is the multioutput kernel in $ \mathbb{R}^{No \times No}$, $k_f$ is a covariance function specifying inter-output similarities and $k_x$ is a covariance function over inputs. In case the inputs are available on a multidimensional cartesian grid, we can further decompose the covariance matrix $\mathbf{K}_x$ as the Kronecker product $\mathbf{K}_{x1} \otimes \cdot\cdot\cdot \otimes \mathbf{K}_{xm}$. Exploiting the Kronecker product property can further improve computational efficiency. 
\subsection{Mean Function: Neural Operator}\label{subsec:NO}
There are many available choices for the mean function in GP, each with its advantages and limitations. For instance, a linear mean function can capture linear trends in the data, while a non-linear mean function, such as a neural network, can handle more complex relationships and nonlinear patterns. By using a mean function that aligns with the characteristics of the problem, the GP can better model the underlying process as it only has to learn the discrepancy between the ground truth and the values computed using the mean function. This results in overall improved predictions. Thus, selecting a suitable mean function in GP is critical in improving the prediction capability, resulting in efficient learning of the underlying function. 
Here in this work, we have used neural operators as the mean function. The core idea of neural operators is to leverage neural networks to learn the underlying function space representation. Neural operators represent a composite function consisting of the encoder, neural network, and decoder. The encoder lifts the dimensionality of the input to a high dimensional latent space. The decoder projects the computed output back to the function space. 
Consider a fixed domain $ D \in (a,b) $ and $x \in D$. For a source term of a PDE $ f(x,t): D \times \mathcal{T}  \mapsto \mathbb{R} $, with initial condition $ u(x,0): D \mapsto \mathbb{R} $ and the boundary condition is $u(\partial D, t) : D \times \mathcal{T} \mapsto \mathbb{R}$. The solution space is $ u(x,t): D \times \mathcal{T}  \mapsto \mathbb{R} $, where $t$ is the time coordinate. For a Banach space with $ \mathcal{A} := \mathcal{C}(D; \mathbb{R}^{d_1})$ and $ \mathcal{U}:= \mathcal{C}(D; \mathbb{R}^{d_2})$, suppose that
\begin{equation}
    \mathcal{D}: \mathcal{A} \times \bm{\theta}_{NN} \mapsto \mathcal{U};  \qquad \bm{\theta}_{NN} = \left\{ \mathcal{W}, \mathbf{b} \right\}
\end{equation}
for arbitrarily any non-linear neural operator $ \mathcal{D}$, where $ \bm{\theta}_{NN} $ is the parameter space of the neural network. If we have access to $N$ number of pairs $\{(a_j, u_j)\}_{j=1}^N$, then we can approximate $\mathcal{D}$ using the pairs of $ N $ data points.
Although the proposed approach can accommodate any neural operator as the mean function, we restrict ourselves to the scenario where Wavelet Neural Operator (WNO) \cite{tripura2023wavelet} is used as the mean function.

WNO exploits the kernel integration proposed in \cite{li2020neural} in combination with wavelet transformation. WNO takes $N$ input functions discretized on a grid and lifts them to a higher dimension using a densely connected layer. The output from the densely connected layer is then fed into the wavelet block, and its dimensions are preserved after the final operations of the wavelet block. 
Mathematically, the wavelet blocks are represented as: 
\begin{equation}
\begin{aligned}
    f_{j+1}(x) = \sigma \left( \left(\mathcal{K}(a(x); \phi \in \bm{\beta}) * v_{j}(x)\right) + b v_{j}(x) \right), \qquad j = 1, \ldots, J,
\end{aligned}
\end{equation} 
where $\sigma$ represents a non-linear activation function, $\mathcal{K}$ denotes the wavelet integral operator, $v_j(x)$ corresponds to the transformed function from the preceding layer, and $b$ signifies a linear transformation. The wavelet coefficients are derived through a discrete wavelet transform (DWT) and an inverse discrete wavelet transform (IDWT) of the mother wavelet. For further details, the interested reader can refer \cite{tripura2023wavelet}. 
\subsection{Likelihood of the data}
Consider $N$ distinct observations with inputs, $\mathbf{Z} = (\bm{x}_1, \ldots, \bm{x}_N,\bm{a}_1, \ldots, \bm{a}_N)^T$, and a corresponding set of outputs, $\mathbf{Y} = (\bm{y}_1, \ldots, \bm{y}_N)^T$ as before. For brevity of representation, we consider the vectorized form of $\mathbf Y$, $\bm q = \left(y_{11}, \ldots, y_{N1},\ldots, \right.$
$\left. y_{1o},\ldots, y_{No} \right)$, where $y_{ij}$ denotes the $j-$th component of the output $\bm y_i$ corresponding to the $i-$th input $\bm x_i$. With this setup, the likelihood can be expressed as follows:
\begin{equation}
    \bm{q} \mid \bm{f}_o,\mathbf{Z}, \bm \theta_m, \bm \theta_k \sim \mathcal{N}(\bm{f}_o(\mathbf{Z}), \sigma_o^2 \mathbf{I}) 
\end{equation}
Given the fact that the mean function is parameterized using the WNO, we set $\bm \theta_m = \bm \theta_{NN}$, and the mean function computed using WNO is represented using $\bm h$. In practice, we work with the Negative log-marginal likelihood (NLML),
% \begin{equation}\label{LML_eq}
% \begin{aligned}
%  \log p(\mathbf{q} \mid \mathbf{X},\bm \theta) & =-\frac{1}{2} (\mathbf{q}-\mathbf{h}(\mathbf{X}; \bm \theta_m))^{\top}\left(\textbf{K}_o(\mathbf{X},\mathbf{X'}; \bm \theta_k)+\sigma_o^2 \mathbf{I}\right)^{-1} (\mathbf{q}-\mathbf{h}(\mathbf{X};\bm \theta_m))-\frac{1}{2} \log \left|\textbf{K}_o(\mathbf{X},\mathbf{X'};\bm \theta_k)+\sigma_o^2 \mathbf{I}\right|+ \text{constant},
% \end{aligned}
% \end{equation}
\begin{equation}\label{LML_eq}
\begin{split}
    \log p(\bm{q} \mid \mathbf{Z},\bm \theta) & =-\frac{1}{2} (\bm{q}-\bm{h}(\mathbf{Z}; \bm \theta_{NN}))^{T}\left(\textbf{K}_o(\mathbf{Z},\mathbf{Z'}; \bm \theta_k)+\sigma_o^2 \mathbf{I}\right)^{-1} (\bm{q}-\bm{h}(\mathbf{Z};\bm \theta_{NN}))\\
    & \quad -\frac{1}{2} \log \left|\textbf{K}_o(\mathbf{Z},\mathbf{Z'};\bm \theta_k)+\sigma_o^2 \mathbf{I}\right|+ \text{constant},
\end{split}
\end{equation}
where the $ \mathbf{I} $ denotes an identity matrix. The unknown hyperparameters $ \bm \theta = \{\bm{\theta}_{NN}, \bm{\theta}_k, \sigma_o\}$ are determined by minimizing the negative log-marginalized likelihood (NLML) as given by eq. \eqref{NLML}. The optimization problem to find the parameter $\bm{\theta}$ is expressed as follows:
\begin{equation}\label{NLML}
    \hat{\bm{\theta}} = \underset{\theta}{\arg\min} \; -\log p(\bm{q} \mid \mathbf{Z}, \bm{\theta})
\end{equation}
Both the parameters of the mean and covariance are optimized simultaneously. Once the training is done the hyperparameter $ \bm \theta$ is obtained. More thorough and rigorous details can be found in Murphy \cite{murphy2022probabilistic} and Rasmussen and Williams \cite{Rasmussen2004}.
An algorithm depicting the steps involved in training the proposed model is shown in Algorithm \ref{alg_gpwno}.
% Algorithm: Training
\begin{algorithm}[ht!]
\caption{Training: NOGaP model }\label{alg_gpwno}
\begin{algorithmic}[1]
\Require{$N$ training samples of input-output pairs $ \{ \bm{z}, \bm{u}(\bm{x})\}$  where spatial points $\bm{x} \in \text{Domain}$, and network hyperparameters.} \\
 Choose a suitable covariance function $k(\bm{z}_i, \bm{z}_j)$  \Comment{Eq. \eqref{Matern Kernel eq}} \\ 
 Set mean function as \textbf{WNO} \\
 Initialize the kernel and mean function's hyperparameters $\bm \theta$
 
\For {Number of iteration = $n$}
\State{Compute the GP output $\hat{\bm{u}}(\bm{x})$ for input $\bm{z}$}
\Comment{Eq. \eqref{p_inference}}
\State{Minimize the negative marginal log-likelihood loss $\mathcal{L}(\bm{u},\hat{\bm{u}})$} \Comment{Eq. \eqref{NLML}}
\State{Compute the gradients of the loss with respect to the model's hyperparameters}
\State{Update the model's hyperparameters using a suitable optimization algorithm}
\EndFor
\State{Save the trained model for later prediction}
\Ensure{Trained NOGaP model with optimized mean function and covariance kernel hyperparameters $ ( \bm \theta)$}
\end{algorithmic}
\end{algorithm}
\subsection{The Predictive Inference}
Since we are working with Gaussian likelihood we can get the predictive distribution in its closed form. We get the following joint distribution for a new data sample $\bm{z^*}$.
\begin{equation}
\begin{pmatrix}
\bm{q} \\
\bm{f}_o^*
\end{pmatrix} \sim \mathcal{N}\left(
\begin{pmatrix}
\bm{h}(\mathbf{Z}; \bm \theta_{NN}) \\
\bm{h}(\bm{z}^*; \bm \theta_{NN})
\end{pmatrix},
\begin{pmatrix}
\mathbf{K}_o(\mathbf{Z}, \mathbf{Z}; \bm{\theta}_k) + \sigma_o^2 \mathbf{I} & \mathbf{K}_o(\mathbf{Z}, \bm{z}^*;\bm{\theta}_k) \\
\mathbf{K}_o( \bm{z}^*,\mathbf{Z};\bm{\theta}_k) & \mathbf{K}_o(\bm{z}^*, \bm{z}^*;\bm{\theta}_k)
\end{pmatrix}
\right)
\end{equation}
Here, the observed training output is denoted by $\bm{q}$, the predictions for the test sample by $\bm{f}_o^*$, and the mean function values for the training samples and the test sample by $\bm{h}(\mathbf{Z}; \bm \theta_{NN})$ and $\bm{h}(\bm{z}^*; \bm \theta_{NN})$.
Computing conditional distribution from the joint distribution is straightforward, refer \cite{bishop2006pattern}. Inference can be done using standard GP formulas for the mean and variance of the predictive distribution.
By conditioning the joint distribution on $\bm{q}$ the posterior predictive distribution of $\bm{f}_o(\bm{z^*})$ can be expressed as:
\begin{subequations}
\begin{align}\label{p_inference}
&p(\bm{f}_o^* \mid \mathbf{Z}, \bm{z}^*, \bm{q}) = \mathcal{N}(\bm{f}_o^* \mid \overline{\bm{m}}^*, \operatorname{cov}(\bm{f}_o^*)), \\
&\overline{\bm{m}}^* = \bm{h}^*(\bm{z}^*; \bm \theta_{NN}) + \mathbf{K}_o(\bm{z}^*, \mathbf{Z}; \bm \theta_k)(\mathbf{K}_o(\mathbf{Z}, \mathbf{Z};\bm \theta_k) + \sigma_o^2 \mathbf{I})^{-1}(\bm{q} - \bm{h}(\mathbf{Z}; \bm \theta_{NN})), \\
&\label{p_inference_cov}\operatorname{cov}(\bm{f}_o^*) = \mathbf{K}_o(\bm{z}^*, \bm{z}^*; \bm \theta_k) - \mathbf{K}_o(\bm{z}^*, \mathbf{Z}; \bm \theta_k)(\mathbf{K}_o(\mathbf{Z}, \mathbf{Z}; \bm \theta_k) + \sigma_o^2 \mathbf{I})^{-1}\mathbf{K}_o(\mathbf{Z}, \bm{z}^*; \bm \theta_k)
\end{align}
\end{subequations}
where, $\overline{\bm{m}}^*$ represents the predictive mean of $\bm{f}_o^*$. The covariance of $\bm{f}_o^*$, denoted as $\operatorname{cov}(\bm{f}_o^*)$, is computed using the Eq. \eqref{p_inference_cov}.
An algorithm depicting the steps involved in inference is shown in Algorithm \ref{alg_gpwno_inf}.

\subsection{Choice of the Covariance function}
One key aspect of any GP model is the choice of the covariance kernel.
Out of a vast range of available kernels, the  Mat\'{e}rn family of kernels is quite popular as it offers a diverse variety of smoothness for modeling signals, and therefore it has been the choice for the numerical examples covered in Section \ref{sec:numerical}. Mathematically, the family of Mat\'{e}rn kernels for $ \mathcal{X} \subset \mathbb{R}^d $, constants $ \alpha > 0 $ and $ h >0 $ can be expressed as
\begin{equation}
    \begin{aligned}
        k_{\alpha, h}\left(\bm{z}, \bm{z}^{\prime}\right)=\frac{1}{2^{\alpha-1} \Gamma(\alpha)}\left(\frac{\sqrt{2 \alpha}\left\|\bm{z}-\bm{z}^{\prime}\right\|}{h}\right)^\alpha K_\alpha\left(\frac{\sqrt{2 \alpha}\left\|\bm{z}-\bm{z}^{\prime}\right\|}{h}\right), \quad \bm{z}, \bm{z}^{\prime} \in \mathcal{X}
    \end{aligned}
\end{equation}
where $\Gamma$ is the gamma function, and $K_\alpha$ is the modified Bessel function of the second kind of order $\alpha$. For specific values of $\alpha$, we get different Mat\'{e}rn kernels namely, Mat\'{e}rn-1/2, Mat\'{e}rn-3/2, and Mat\'{e}rn-5/2. The following equation shows the mathematical expression for the Mat\'{e}rn-5/2:
\begin{subequations}\label{Matern Kernel eq}
\begin{align}
& k_{5 / 2, h}\left(\bm{z}, \bm{z}^{\prime}\right)=\left(1+\frac{\sqrt{5}\left\|\bm{z}-\bm{z}^{\prime}\right\|}{h}+\frac{5\left\|\bm{z}-\bm{z}^{\prime}\right\|^2}{3 h^2}\right) \exp \left(-\frac{\sqrt{5}\left\|\bm{z}-\bm{z}^{\prime}\right\|}{h}\right).
\end{align}
\end{subequations}

%Inference algorithm
\begin{algorithm}[ht!]
\caption{Inference: NOGaP model}\label{alg_gpwno_inf}
\begin{algorithmic}[1]
\Require{$N^*$ test samples $ \bm{z} =\{\bm{x},\bm{a}(\bm{x})\}$ where spatial points $\bm{x} \in \text{Domain}$, and trained network hyperparameters $\bm \theta$.} 
\For{each test sample}
    \State{Compute the GP output \(\hat{\bm{u}}(\bm{x})\) for input $\bm{z}$ using the trained GP model.}
    \Comment{Refer to Eq. \eqref{p_inference}}
    \State{Store the prediction \(\hat{\bm{u}}(\bm{x})\) corresponding to the input $\bm{z}$.}
\EndFor
\Ensure{Predictions \(\hat{\bm{u}}(\bm{x})\) corresponding to inputs $\bm{z}$ for all test samples.}
\end{algorithmic}
\end{algorithm}

\begin{figure}[ht!] 
	\centering
	\includegraphics[scale=0.5]{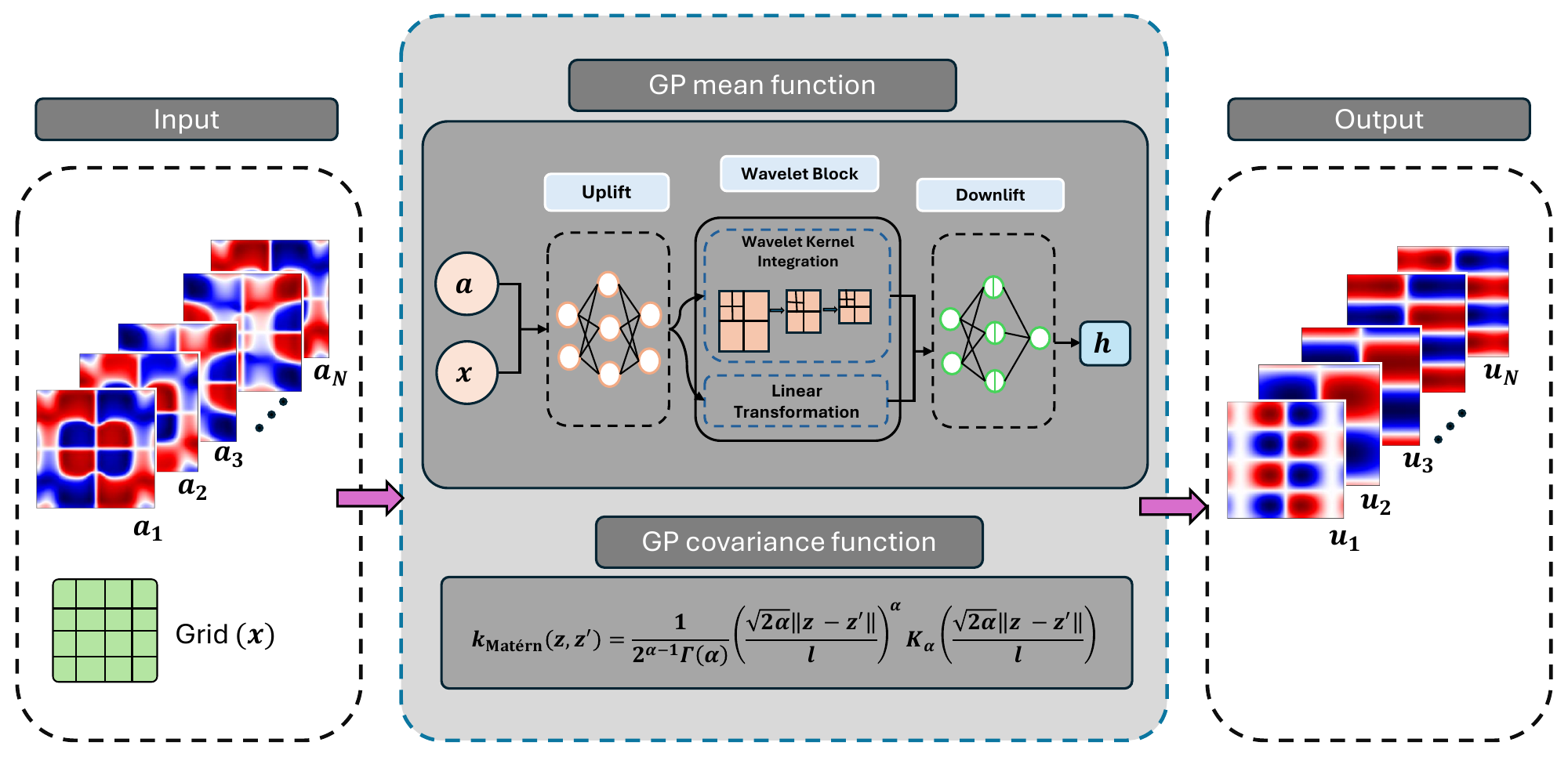}
	\caption{\textbf{Schematic of NOGaP:}  The illustration depicts the NOGaP framework, featuring a Mean and Covariance block. The computation of the mean function is facilitated by the Wavelet Neural Operator (WNO) block. NOGaP operates by utilizing N-training samples comprising input-output pairs alongside the grid. The corresponding N outputs encompass both the mean of the predicted samples and the associated standard deviation.}
	\label{fig: Schematics NOGaP}
\end{figure}

\section{Numerical examples}\label{sec:numerical}
This section explores a variety of case studies ranging from fluid dynamics, gas dynamics, and phase-field modeling on which the proposed NOGaP framework is tested for its efficacy. The examples include (a) 1D Burger's equation, (b) 1D wave-advection equation, (c) 2D Darcy flow equation with a notch in the triangular domain, and (d) 2D non-homogeneous Poisson equation. The architecture of NOGaP consists of two key blocks: mean function and covariance function. The Daubechies family of mother wavelets is used for the mean function block with the GeLU which is differentiable at all points \cite{hendrycks2023gaussian}. The information related to the architecture of the framework is summarised in \autoref{table: Mean_f_archi}.
The framework's performance is summarised in the \autoref{table_accuracy}, displaying a comparison of relative Mean Square Errors (MSE) and the comparison with WNO, GP (zero mean) and Bayesian WNO (B-WNO)  for various examples.
The results obtained from NOGaP demonstrate that the proposed framework shows improved accuracy and can learn the desired mapping between input and output space effectively. The proposed NOGaP framework also provides the uncertainty estimates associated with the predictions. Further details of individual cases are illustrated in the following subsection.

\begin{table}[ht!]
    \centering
    \caption{NOGaP's architecture for each numerical example; Mat\'{e}rn kernel is used for covariance function in all cases.}
    \label{table: Mean_f_archi}
    \begin{tabular}{p{4cm}cccccccc}
    \hline
    \multirow{2}{*}{Examples} & \multicolumn{2}{c}{Data size} & \multirow{2}{*}{Wavelet} & \multicolumn{5}{c}{NN dimensions (mean function)} \\ \cline{2-3}\cline{6-9}
     & Training & Testing & & & FNN1 & FNN2 & CNN & level  \\
    \hline
    Burgers$^{(\ref{sec:bur_eq})}$ & 1000& 50 & db6 &  & 64 & 128 & 4 & 8 \\
    Advection$^{(\ref{sec:AD_eq})}$ & 1000 & 50 & db8 &  & 96 & 128 & 4 & 4 \\
    Darcy (triangular)$^{(\ref{sec:darcy_td})}$ & 500 & 50 & db6 &  & 32 & 192 & 4 & 6  \\
    Poisson $^{(\ref{sec:poisson})}$ & 500 & 50 & db4 &  & 64 & 132 & 4 & 4\\
    \hline
    \end{tabular}
\end{table}

\subsection{1D Burger Equation}\label{sec:bur_eq}
\begin{figure}[ht!]
	\centering
	\includegraphics[width=\textwidth]{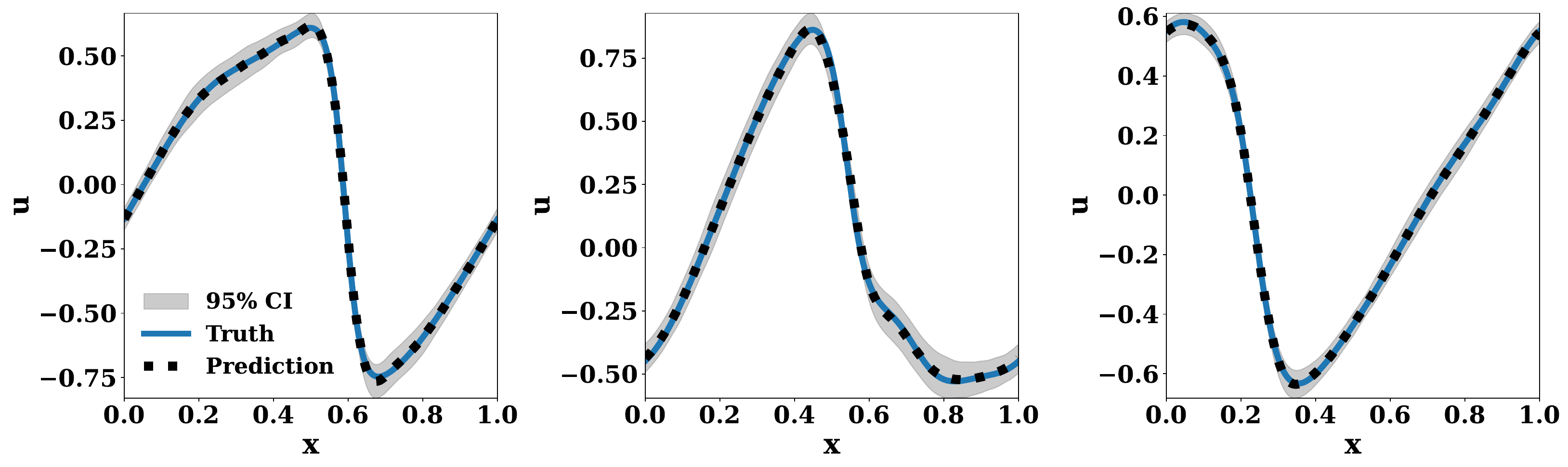}
	\caption{\textbf{1D Burger equation with periodic boundary condition.} This figure shows the ground truth and predicted solution for different initial conditions along with the $95\%$ confidence interval (grey region) associated with each prediction. The results clearly show that NOGaP can make excellent predictions for this example.}
	\label{fig_burger1d_2}
\end{figure}

\begin{figure}[ht!]
	\centering
	\includegraphics[width=\textwidth]{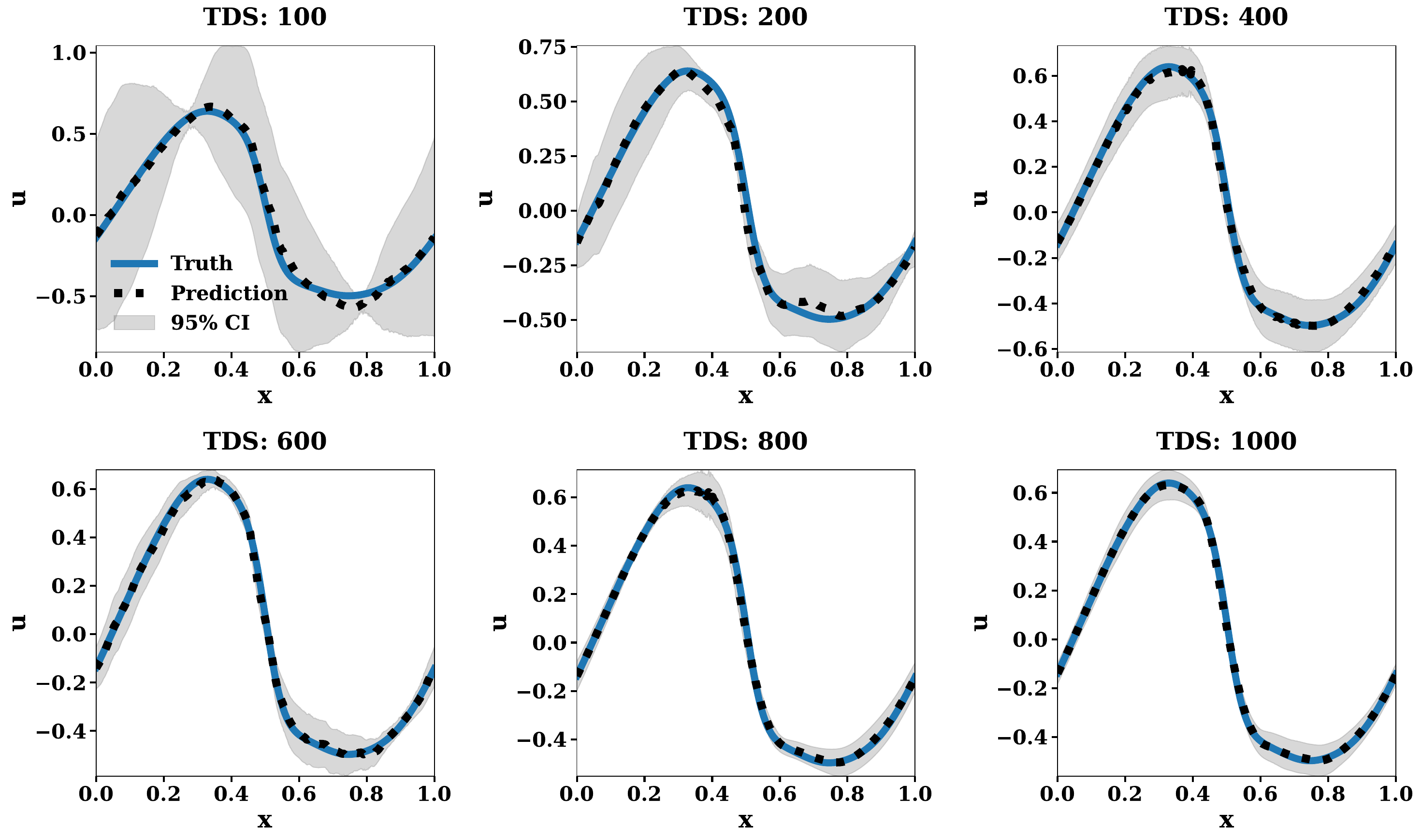}
	\caption{\textbf{NOGaP predictions for 1D Burgers' equation for different sizes of training data samples (TDS).} This figure shows the ground truth and the predicted solution along with the $95\%$ confidence interval (CI). It can be observed that the predictive uncertainty decreases with an increase in training sample sizes, thus reducing the CI. It is the expected behavior in GPR.}
	\label{1D_Burger_TDS}
\end{figure}

The 1D Burgers equation is a well-known partial differential equation that models one-dimensional flows in various fields, including fluid mechanics, gas dynamics, and traffic flow. One of the interesting features of this equation is that it exhibits both diffusive and advective behavior, making it a useful tool for studying a wide range of phenomena. The 1D Burger equation with periodic boundary is mathematically expressed as,

\begin{equation}\label{eq_BS}
\begin{aligned}
\partial_{t} u(x, t) + 0.5\partial_{x} u^{2}(x,t) &= \nu \partial_{x x} u(x,t), & & x \in(0,1), t \in(0,1] \\
u(x = 0,t) &= u(x= 1,t), & & x \in(0,1), t \in(0,1] \\
u(x,0) &= u_{0}(x), & & x \in(0,1).
\end{aligned}
\end{equation}
$\nu$ here is the viscosity of the flow, which is positive and greater than zero, and $u_{0}(x) $ is the initial condition. For generating the initial conditions $u_{0}(x)$, a Gaussian random field is used where $u_{0}(x) \sim \mathcal{N}(0,625(- \Delta + 25I)^{-2}.$ The aim is to learn the mapping from input and output function spaces for the 1D Burger's PDE that is $ u(x, t = 0) \mapsto u(x, t = 1) $. We consider $\nu = 0.1$ and a spatial resolution of 512. The dataset is taken from Ref. \cite{li2020fourier}.  
\textbf{Results:} The predictions for three different initial conditions are presented in \autoref{fig_burger1d_2}. The relative MSE values of the predictions are mentioned in \autoref{table_accuracy}, revealing that the proposed framework exhibits favorable performance for the 1D Burger's problem. Notably, it provides an improvement over the 
standalone WNO, GPR (zero mean) and B-WNO predictions. A visual inspection of the plot confirms the agreement between the ground truth obtained from the numerical solver and the predictions generated by the NOGaP framework. Additionally, the plot includes the corresponding 95\% confidence intervals, providing a measure of uncertainty associated with the predictions. Further, a case study is carried out by considering different training sample sizes (see \autoref{1D_Burger_TDS}). The mean response for all the cases is found to have an excellent match with the ground truth. The predictive uncertainty, on the other hand, reduces with an increase in the number of training samples. This indicates a reduction in the predictive uncertainty with an increase in training samples.

\subsection{1D Wave Advection Equation}\label{sec:AD_eq}
The wave advection equation is a type of hyperbolic partial differential equation (PDE) that is commonly used in physics and engineering to describe the movement of a scalar under a known velocity field. The equation is often used to model the behavior of waves in various media, such as sound waves in air or water waves in the ocean. In this example, periodic boundary conditions are applied to simulate the behavior of the medium over time.
When the wave advection equation is considered with periodic boundary conditions, it can be expressed mathematically as:

\begin{equation}
    \begin{aligned}
    \partial_t u(x,t) + \nu \partial_x u(x,t) &= 0, & & x \in (0,1), t\in (0,1) \\
    u(x - \pi) &= u(x+\pi), & & x \in (0,1).
    \end{aligned}
\end{equation}
Here $\nu $ is positive and greater than zero,$u$ representing the speed of the flow and the boundary condition is $2\pi$ periodic. The initial condition is chosen as,
\begin{equation}\label{wave_init}
        u(x,0) = h1_{ \left\{c- \frac{\omega}{2}, c+\frac{\omega}{2} \right\} } + \sqrt{ \max(h^2 - (a(x-c))^2,0) }.
\end{equation}
here the variable $\omega$ represents the width and $h$ represents the height of the square wave. The wave is centred around at $x = c$ and the values of \{c,$\omega$,$h$\} are selected from $[0.3,0.7]\times [0.3,0.6] \times [1,2]$. For this example, the value of $\nu $ is set to 1 and we are learning the mapping from the initial condition to the final condition at $ t = 0.5$ that is $ u(x,t=0) \mapsto u(x,t=0.5)$. The spatial resolution grid is 40 and we have considered 1000 instances of training samples. 

\begin{figure}[ht!]
    \centering
    \includegraphics[width=\textwidth]{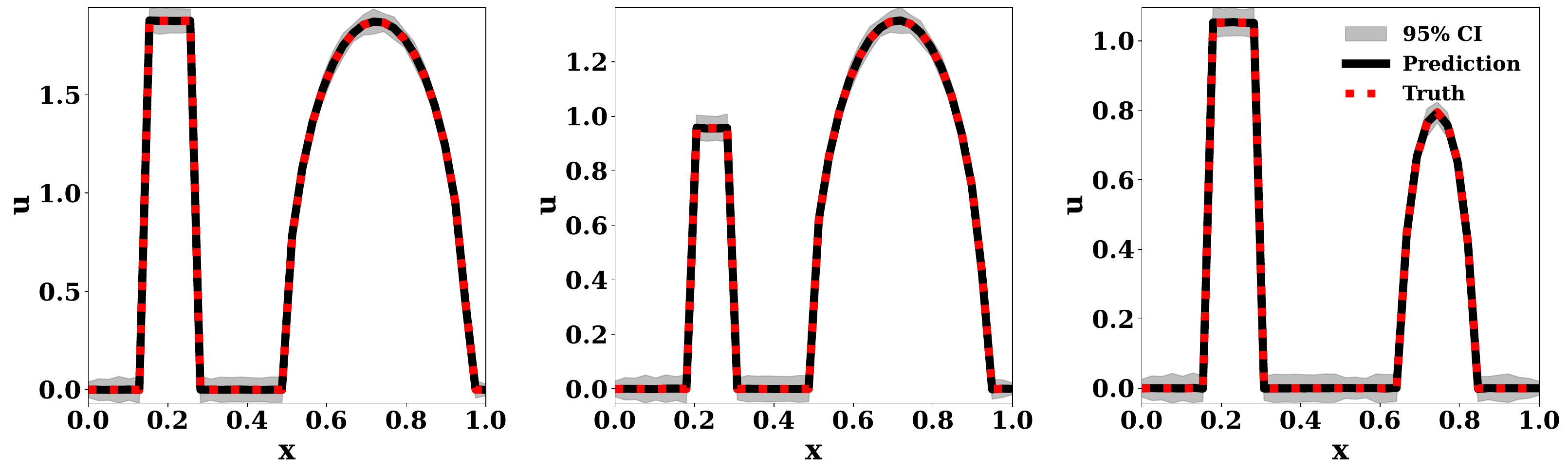}
    \caption{\textbf{1D Wave advection equation with periodic boundary condition}. The plot illustrates the predictions of the NOGaP model, which effectively learns the mapping between input and output function spaces.  The predicted solutions for three different initial conditions obtained by the proposed NOGaP framework for $u(x,t)$ at $t=0.5$ are plotted, along with a 95 \% confidence interval.}
 \label{advection_1d}
\end{figure}

\begin{figure}[ht!]
	\centering
	\includegraphics[width=\textwidth]{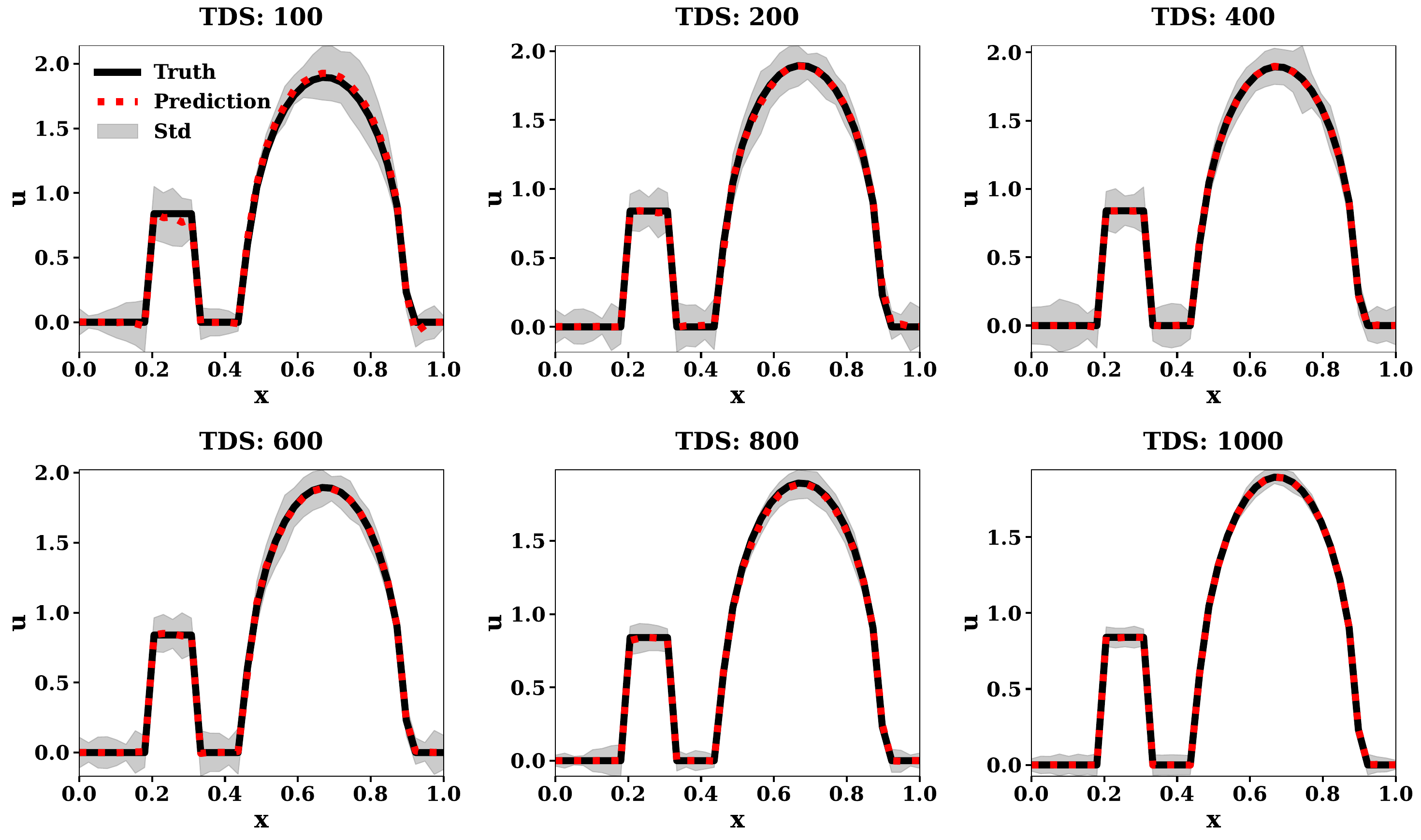}
	\caption{\textbf{NOGaP predictions for 1D Wave advection equation for different training sample sizes.} The plot shows the ground truth and predicted solution along with the $95\%$ CI. It can be observed that the predictive uncertainty decreases with an increasing number of training sample sizes.}
	\label{1D_advection_TDS}
\end{figure}

\textbf{Results:} \autoref{advection_1d} depicts the predictions obtained from the NOGaP for different initial conditions. The \autoref{table_accuracy} shows the relative MSE value from NOGaP and other frameworks. The RMSE value for NOGaP is better compared to WNO, GP (zero mean) and B-WNO. Similar to the previous example it can be observed that the NOGaP model can make predictions effectively for this example. Further, a case study is carried out by considering different training sample sizes (see \autoref{1D_advection_TDS}). The mean response for all the cases is found to have an excellent match with the ground truth. The predictive uncertainty, on the other hand, reduces with an increase in the number of training samples. This indicates a reduction in the predictive uncertainty with an increase in training samples.

\begin{table}[ht]
	\centering
	\caption{Relative Mean Square Error \textbf{(RMSE)} between the ground truth and the predicted results.}
	\label{table_accuracy}
	\begin{threeparttable}
	\begin{tabular}{ccccccc:c}
		\hline
		\multirow{2}{*}{} & \multicolumn{7}{c}{\textbf{PDEs}}\\ \cline{2-8}
		& Burger & Wave Advection & Darcy (triangular) & Poisson \\
		\hline
		\textbf{NOGaP} & $\approx$ 4.097 $\pm$0.307 \% & $\approx$ 0.232 $\pm$ 0.01 \% & $\approx$ 5.798 $\pm$ 0.407\% & $\approx$ 0.278 $\pm$ 0.033 \% \\
		\textbf{WNO} & $\approx$ 6.556 $\pm$ 0.794
         \% & $\approx$ 0.599 $\pm$ 0.127 \% & $\approx$ 5.589  $\pm$ 0.715  \% & $\approx$ 3.710 $\pm$ 0.593  \%    \\
        \textbf{B-WNO} & $\approx$ 21.434 $\pm$ 0.784 \% & Did not converge & $\approx$ 11.714 $\pm$ 0.987\% & $\approx$ 9.042 $\pm$ 0.484 \% \\
        \textbf{GP (zero mean)} & $\approx$ 4.581 $\pm$ 0.111 \% & $\approx$ 5.186 $\pm$ 0.509\% & $\approx$ 11.004 $\pm$ 0.461\% & $\approx$ 1.503 $\pm$ 0.245 \% \\
        \hline
        % *where DNC (Did not converge) & &  &  & 
	\end{tabular}
	\end{threeparttable}
\end{table}

\subsection{2D Darcy flow equation with a notch in triangular domain}\label{sec:darcy_td}
The 2D Darcy flow equation is a popular tool utilized for modeling the flow of gases or liquids through a porous medium. This equation is particularly useful in fields such as environmental engineering, geology, and hydrology, where understanding fluid dynamics in porous media is essential. Without the time component, the 2D Darcy flow equation can be represented as a second-order nonlinear elliptic partial differential equation (PDE). The equation is given as,

\begin{equation}
 \begin{aligned}
     -\nabla \cdot (a(x,y)\nabla u(x,y) &= f(x,y), & & x,y \in(0,\mathbb{R}) \\
     u(x,y) &= u_{0}(x,y), & & x,y \in \partial(0,\mathbb{R}) 
 \end{aligned}
\end{equation}
here we have $u(x,y) = u_{0}(x,y)$ is the initial condition, $a(x,y)$ is the permeability field, $u(x,y)$ is the pressure, $f(x,y)$ is a source function, and $\nabla u(x,y)$ is the pressure gradient. For this example, the setup and dataset are taken from Ref. \cite{li2020fourier}. In this example, a complex boundary condition is considered that is a triangular domain with a notch. The dataset is generated using Gaussian Process (GP) as follows:
\begin{equation}
    \begin{aligned}
    u(x) \sim \mathcal{GP}(0,k(x,x'), \\
    k(x,x') = \exp \left(-\frac{(x-x')^2}{2l^2} \right), & & l=0.2, & & x,x' \in [0,1]
    \end{aligned}
\end{equation}
A notch is introduced in the flow medium with the triangular geometry. The forcing function $f(x,y)$ is set to -1 and the permeability field $a(x,y)$ is set to 0.1. The objective is to learn the mapping from the input (boundary conditions) and output (pressure field) function spaces that are $ u(x,y)|_{\partial \omega} \mapsto u(x,y)$. The spatial resolution is taken to be $ 26 \times 26 $.

\begin{figure}[ht!]
	\centering
        \includegraphics[width=\textwidth]{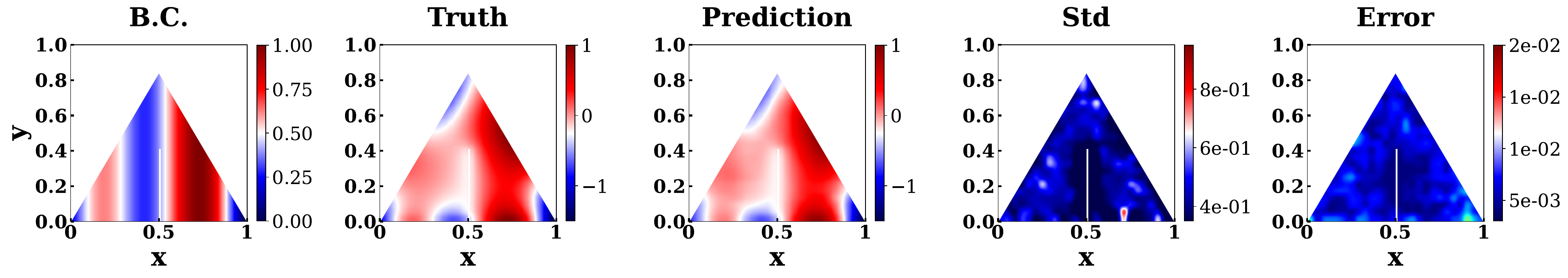}
	\caption{\textbf{2D Darcy flow with a notch in the triangular domain}. The figure shows the prediction obtained from the proposed NOGaP framework for one of the representative test samples. The plot shows the boundary condition, the ground truth, the mean and standard deviation of the prediction and the corresponding error.}
	\label{fig_darcy2d_notch}
\end{figure}

\textbf{Results:} The \autoref{fig_darcy2d_notch} shows the prediction obtained from the proposed NOGaP model. It displays predictions corresponding to one of the boundary conditions. The corresponding ground truth is obtained from the numerical solver and the mean predictions from the NOGaP. The plot suggests that the proposed framework can make accurate predictions and is able to approximate the true solution. From the \autoref{table_accuracy} we can observe, that there isn't much difference between the RMSE value for both NOGaP and WNO but since the former provides an uncertainty measure over the predictions it becomes more useful and reliable. However, the predictions are significantly improved when compared with GP (zero mean) and B-WNO. Through visual inspection, one can hardly find any discernable difference between the ground truth and the predictions.

% \begin{figure}
%     \centering
% \label{fig_darcy2d_notch1}
% \begin{subfigure}{\textwidth}
% \centering
% \includegraphics[scale=0.25]{Row_darcy_notch_first02.pdf}
% \end{subfigure}
% \begin{subfigure}{\textwidth}
% \centering
% \includegraphics[scale=0.25]{Row_darcy_notch_second02.pdf} 
% \end{subfigure}
% \caption{\textbf{2D Darcy flow with a notch in the triangular domain}. The figure shows the prediction obtained from the proposed NOGaP framework on one of the representative test samples. It is evident from the plot that the proposed framework can predict the solution with comparable accuracy with the standalone WNO for different boundary conditions.}
% \label{fig_darcy2d_notch}
% \end{figure}

\subsection{2D Non-homogeneous Poisson's Equation}\label{sec:poisson}
The non-homogeneous Poisson equation is an elliptic partial differential equation (PDE) widely applicable in diverse physical scenarios. It serves as the fundamental mathematical model for various problems, such as the steady-state heat diffusion and the electric field generated by a specified electric charge. The Poisson equation, incorporating a source term \( f(x, y) \) and periodic boundary conditions, is mathematically expressed as:
\begin{equation}
 \begin{aligned} \label{poisson_eq}
     \partial_{xx}u + \partial_{yy}u = f(x, y), \quad x \in [-1, 1], \quad y \in [-1, 1] \\
      u(x = -1, y) = u(x = 1, y) = u(x, y = -1) = u(x, y = 1) = 0 
 \end{aligned}
\end{equation}
 where, $ f(x,y) $ is the source function, $ u(x,y) $ is the solution of the equation over the 2D domain. The objective is to learn the mapping between the input and the output space, that is $  f(x,y) \mapsto u(x,y) $.
 To establish the ground truth solution, we opt for an analytical solution represented as \(u(x, y) = \alpha \sin(\pi x)(1 + \cos(\pi y)) + \beta \sin(2\pi x)(1 - \cos(2\pi y))\). For generating input data, we employ an analytical form for the source function, defined as \(f(x, y) = 16\beta\pi^2 (\cos(4\pi y) \sin(4\pi x) + \sin(4\pi x)(\cos(4\pi y) - 1)) - \alpha\pi^2 (\cos(\pi y) \sin(\pi x) + \sin(\pi x)(\cos(\pi y) + 1))\). This expression for \(f(x, y)\) is derived by substituting the analytical expression for \(u(x, y) = \alpha \sin(\pi x)(1 + \cos(\pi y)) + \beta \sin(2\pi x)(1 - \cos(2\pi y))\) into \autoref{poisson_eq}. Consequently, the solution implicitly satisfies the boundary conditions. The training instances are generated by varying the parameters \(\alpha\) and \(\beta\) such that \(\alpha \sim \text{Unif}(-2, 2)\) and \(\beta \sim \text{Unif}(-2, 2)\). The spatial resolution of $33 \times 33$ is taken for this example. 

\begin{figure}[ht!]
	\centering
	\includegraphics[width=\textwidth]{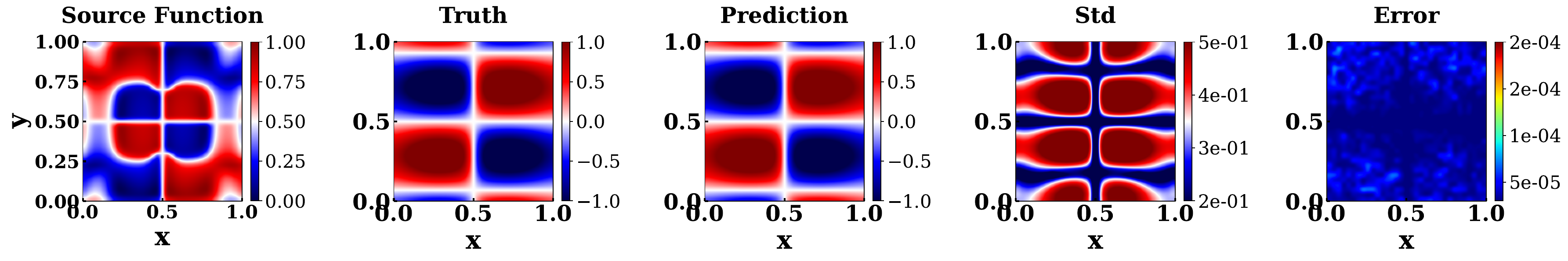}
	\caption{\textbf{2D Non-homogeneous Poisson's equation }.The figure shows the prediction obtained from the proposed NOGaP framework for one of the representative test samples. The plot shows the source function, the ground truth, the mean and standard deviation of the prediction and the corresponding error.}
	\label{fig_poisson}
\end{figure}

\textbf{Results:} The \autoref{fig_poisson} shows the predictions obtained from the proposed NOGaP model. It displays predictions corresponding to one of the source functions. The plot consists of the ground truth obtained from the numerical solver and the mean predictions from the NOGaP framework. The plot suggests that the proposed framework can make predictions effectively. From \autoref{table_accuracy}, we can observe that the Relative Mean Square Error \textbf{(RMSE)} value is significantly improved over \textbf{WNO}. Thus, we can conclude that \textbf{NOGaP} is an improvement over the standalone \textbf{WNO}, \textbf{GPR} (zero mean) and \textbf{B-WNO}. Moreover, visually, the predicted and true solutions are almost indiscernible.

% \begin{figure}
% \centering
% \begin{subfigure}{\textwidth}
% \centering
% \includegraphics[scale=0.25]{Row_poisson_first_02.pdf}
% \end{subfigure}
% \begin{subfigure}{\textwidth}
% \centering
% \includegraphics[scale=0.25]{Row_poisson_second_02.pdf} 
% \end{subfigure}
% \caption{\textbf{2D Non-homogeneous Poisson's equation}. The figure shows the prediction obtained from the proposed NOGaP framework on one of the representative test samples. It is evident from the plot that the proposed framework can predict the solution effectively with improved accuracy over the standalone WNO for different source functions.}
% \label{fig_poisson}
% \end{figure}

\section{Conclusion}\label{sec: conclusion}
This paper introduced a novel framework called NOGaP for effectively learning solutions to non-linear PDEs. By harnessing the capabilities of GP, the proposed framework not only provides accurate predictions but also provides a measure of uncertainty. By incorporating mean functions to be WNO, the NOGaP framework demonstrates effective performance in learning the solutions of non-linear PDEs.
Compared to a standalone WNO, GPR (zero mean) and B-WNO, the proposed framework offers distinct advantages in terms of improved prediction capabilities.

To validate the effectiveness of the framework, four numerical examples were considered, and the predictions were compared against the ground truth obtained using numerical solvers. The results demonstrate that the NOGaP framework achieves improved accuracy, with the predictions closely matching the ground truth solutions. The following observation has been made from the results obtained: 
\begin{itemize}
    \item The predictions obtained from the NOGaP framework demonstrate a strong agreement with the ground truth, providing a good approximation of the true solutions.
    \item The selection of a suitable mean function poses a significant challenge in constructing effective frameworks involving GP. However, by incorporating the Wavelet Neural Operator (WNO) as the mean function, the NOGaP framework achieves improved accuracy over standalone WNO, GP (zero mean) and B-WNO.
    \item Additionally, the NOGaP framework is convenient to implement and applicable to a wide variety of examples, making it a versatile tool for solving non-linear partial differential equations.
\end{itemize}

The results suggest that the proposed NOGaP can make predictions effectively and efficiently. We have used vanilla kernels as the choice of covariance function, but it will be interesting to note the effect of other possible kernels on the proposed NOGaP framework. Working with GPs is computationally expensive, especially with large and high dimensional datasets, thus further study can be done to bring down the computational cost.

\section*{Acknowledgements}
SK acknowledges the support received from the Ministry of Education (MoE) in the form of Research Fellowship. RN acknowledge the financial support received from the Science and Engineering Research Board (SERB), India via grant no. SRG/2022/001410. SC acknowledge the financial support received from the Ministry of Port and Shipping via letter no. ST-14011/74/MT (356529). Faculty Seed grants received from IIT Delhi are also acknowledged.

\section*{Code availability}
On acceptance, all the source codes to reproduce the results in this study will be made available to the public on GitHub by the corresponding author.

% Bibliography
% ~~~~~~~~~~~~~~~~~~~~~~~~~~~~~~~~~~~~~~~~~~
% \bibliographystyle{unsrt}  
% \bibliography{bibliography}  

\end{document}